\documentclass[letterpaper]{article} 
\PassOptionsToPackage{table}{xcolor}
\usepackage{aaai2027}  
\usepackage[hyphens]{url}  
\usepackage{graphicx} 
\usepackage{natbib}  
\usepackage{caption} 
\usepackage{algorithm}
\usepackage{algorithmic}

\usepackage{newfloat}
\usepackage{listings}
\DeclareCaptionStyle{ruled}{labelfont=normalfont,labelsep=colon,strut=off} 
\floatstyle{ruled}
\newfloat{listing}{tb}{lst}{}
\floatname{listing}{Listing}

\usepackage{booktabs}
\usepackage{xspace}
\usepackage{amsmath}
\usepackage{amssymb}
\usepackage{mathtools}
\usepackage{amsthm}
\usepackage{algorithm}
\usepackage{algorithmic}
\usepackage{multirow}
\usepackage{pifont}
\usepackage[most]{tcolorbox}

\newcommand{\dataset}{\textsc{\small{StateMAS}}\xspace}
\newcommand{\system}{\textsc{\small{MARS}}\xspace}
\newcommand{\bestscore}[1]{\textbf{#1}}
\newcommand{\INPUT}{\STATE \textbf{Input: }}
\newcommand{\OUTPUT}{\STATE \textbf{Output: }}

\title{Autonomous Repair for Multi-Agent Systems via Monte-Carlo Tree Search}
\author{
    Hanxiao Lu, Tianyi Zhang
}
\affiliations{
    Purdue University\\
    lu525@purdue.edu, tianyi@purdue.edu
}

\begin{document}

\maketitle

\begin{abstract}
Multi-agent systems (MAS) are increasingly deployed to solve complex tasks. In case of incorrect or unsatisfactory outputs, users have to manually locate agent mistakes by inspecting agent trajectories (i.e., {\em failure attribution}) and provide feedback to refine the outputs (i.e., {\em repair}).  
Despite some recent work in MAS failure attribution, automated mechanisms to recover from such mistakes remain largely unexplored. 
To bridge this gap, we propose \system, a search-based framework that formulates MAS repair as a Monte Carlo Tree Search (MCTS) process and navigates the vast space of potential repairs via diagnosis-guided expansion with taxonomy-augmented evaluation. Unlike standard MCTS, which evaluates a complete simulation via full rollout, \system evaluates the agent trajectory using partial rollout to reduce token consumption. Furthermore, we introduce \dataset, a large-scale MAS repair benchmark with 1,310 replayable multi-agent failure trajectories spanning four types of agent architectures and four LLM backbones. Experiments on \dataset demonstrate that  \system consistently outperforms state-of-the-art methods, achieving an absolute improvement from 3.0\% to 12.1\% across all settings, while maintaining a comparable token consumption cost. The ablation study further confirms that taxonomy-augmented evaluation and diagnosis-guided expansion are critical to achieving these performance gains.
\end{abstract}


\section{Introduction}

Multi-agent systems (MAS) have received significant attention and are increasingly deployed to solve complex tasks~\cite{wu2025webwalker, zhang2025webpilot,qian2024chatdev, hong2023metagpt}. 
However, when a MAS produces an incorrect or unsatisfactory output, users have to manually inspect its agent trajectories to locate mistakes (i.e., \textit{failure attribution}) and provide feedback to refine the output (i.e, \textit{repair}). This manual process is tedious and time-consuming.





To reduce this manual burden, recent research has focused on automating failure attribution in MAS~\cite{zhang2025agent, cemri2025multi, zhang2026agentforesight, zhang2026agentracer}. Although identifying the root cause is a crucial first step, automated MAS repair remains largely unexplored. 
To the best of our knowledge, DoVer~\cite{ma2025dover} is the only work for automated MAS repair. Specifically, it segments a MAS trajectory into trails using re-plan steps as segmentation points, identifies suspicious steps, refines agent messages, and re-executes the agentic system from there. Although DoVer represents a pioneering step toward automated MAS repair, its sequential exploration strategy could limit the search space in high token costs. Furthermore, DoVer focuses on repair at the orchestrator level by directly editing the orchestrator's plan or its message to a sub-agent, which limits its applicability to agentic systems without a centralized orchestrator. 

To overcome these limitations,  we propose \system, which performs Monte Carlo Tree Search (MCTS) to explore the vast space of potential repairs to an MAS trajectory. Instead of committing to a single, monolithic trajectory regeneration, \system systematically explores alternative repair solutions by iteratively applying specialized \textit{Rollback}, \textit{Guided Repair}, and \textit{Continuation} actions. This process, termed diagnosis-guided expansion, utilizes partial rollout (i.e., re-executing the agentic system up to $k$ steps) rather than full rollouts (i.e., re-executing the system to completion). This design significantly reduces token consumption while providing fine-grained control over trajectory exploration. However, since partial rollouts do not produce complete trajectories, their quality cannot be assessed using the final task outcome. To address this challenge, we introduce a novel taxonomy-augmented evaluation mechanism that characterizes common failure patterns and generates dynamic reward signals for MCTS, enabling informed search decisions even before a complete trajectory is available.

Evaluating MAS repair methods requires datasets that move beyond static, non-replayable agent trajectories from MAS  failure attribution benchmarks~\cite{zhang2025agent, cemri2025multi, zhang2026agentforesight, zhang2026agentracer}. To this end, we introduce \dataset, a large-scale benchmark with 1310 replayable MAS failure trajectories spanning across four types of agent architectures and four LLM backbones on two popular agent benchmarks, GAIA~\cite{mialon2023gaia} and AssistantBench~\cite{yoran2024assistantbench}. Specifically, \dataset captures the complete system state at every agent step and provides interfaces to deterministically replay the agent execution based on a logged trajectory and roll back the execution to any arbitrary step for intervention. 

We evaluate \system on \dataset against DoVer~\cite{ma2025dover} and two alternative repair methods based on Reflexion~\cite{shinn2023reflexion} and ReAct~\cite{yao2023react}. The results show that \system consistently outperforms all baselines in all settings. Specifically, compared to DoVer, \system achieved absolute improvement from 8.5\% to 10.3\% on GAIA and from 6.1\% to 12.2\% on AssistantBench across four LLM backbones. Furthermore, \system's performance improvement is generalizable when repairing task failures of MAS with different architectures, ranging from 3.0 \% to 12.1\% compared to the strongest baseline. Notably, with the partial rollout design, these improvements are not achieved by inflating computational cost. \system maintains token consumption comparable to the most economical baseline with only 5.6\% overhead, while consuming up to 59.1\% fewer tokens than the most expensive baseline. Finally, the ablation study confirms that these performance gains are driven by the critical synergy between our specialized repair actions and the taxonomy-augmented evaluation mechanism.

\section{Problem Formulation}
\label{sec: problem formulation}



We formalize a multi-agent system as a tuple $\mathcal{M} = \langle \mathcal{N}, \mathcal{S}, \mathcal{A}, \mathcal{T} \rangle$, where:
\begin{itemize}
    \item $\mathcal{N} = \{1, \dots, n\}$ is the set of $n$ distinct agents.
    \item $\mathcal{S}$ represents the global state space of MAS. Each state $s_t \in \mathcal{S}$ is a a replayable snapshot of the system after step $t$. It contains the conversation history and the runtime information required to resume execution, such as agent activation status, orchestration control variables, per-agent memory, and tool outputs.
    \item $\mathcal{A}$ denotes the joint action space of all agents, such as generating a natural language response and calling a tool.
    \item $\mathcal{T}$ is the state transition function governing system evolution $\mathcal{S} \times \mathcal{A} \to \mathcal{S}$. This function specifies the interaction between agents. It defines how an action $a_t$ by an agent updates the global context to $s_{t+1}$ and determines which agent $i \in \mathcal{N}$ is activated next. 
\end{itemize}

The multi-agent system trajectory is defined as $\tau = (s_0, a_0, s_1, a_1, \dots, s_T)$. 
In this work, we assume access to an initial trajectory $\tau_{fail}$ that leads to a task failure. Following prior work~\cite{zhang2025agent, cemri2025multi}, we define a task failure as any MAS run that does not achieve the intended task objective. Such failures may stem from various reasons, e.g., incorrect agent reasoning, task misunderstanding, tool call errors, etc. Such failures may also arise from multiple errors introduced by different agents at different stages of the trajectory. 

A repair intervention is defined as $\delta = (t, \phi)$, where $t$ represents the rollback step and $\phi$ represents the repair operation. Applying an intervention $\delta$ reverts the system state back to $t$, performs the repair operation $\phi$, and resumes the system execution from there, which produces a refined trajectory $\tau'$ with a new final state $s_T'$. 
Since there may be multiple errors in the original trajectory and the repair intervention may also lead to new errors, a single intervention is often insufficient. Thus, the goal is to search for a \textit{sequence} of interventions $\Delta^* = \{\delta_1, \delta_2, \dots, \delta_k\}$ that iteratively fix any potential errors and guide the system to a refined output. During the search process, we do not assume access to an oracle (e.g., a human evaluator, a ground-truth result, a test suite, etc.) that can be repeatedly queried to assess the final outcome of a trajectory. In practice, such an oracle is often unavailable or expensive to query at inference time.


\begin{figure*}[htbp]
    \centering
    \includegraphics[width=0.98\textwidth]{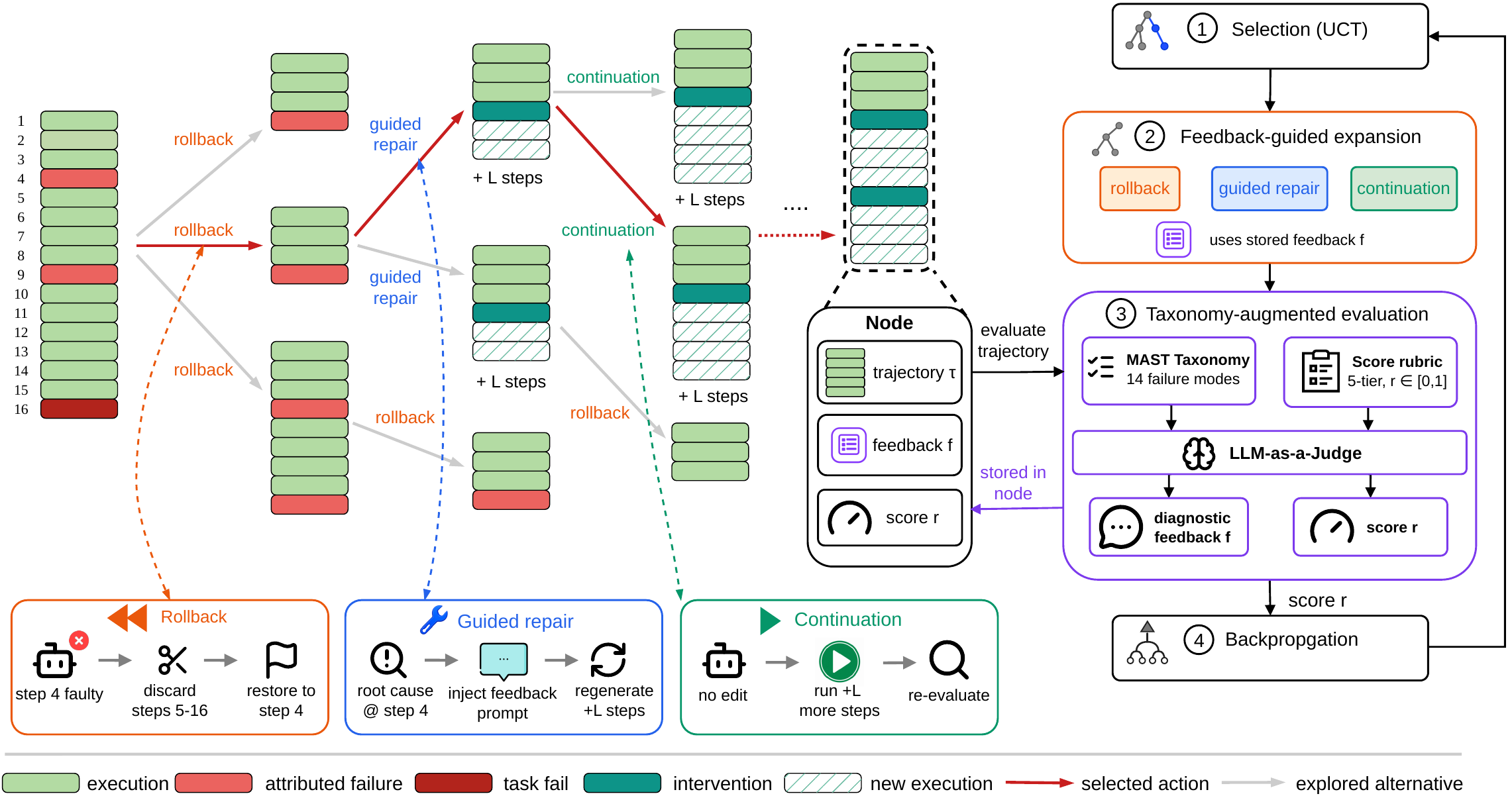}
    \caption{The overview framework of \system.}
    \label{fig: system overview}
\end{figure*}

\section{Method}



To efficiently navigate the vast space of possible repair interventions, we propose a \underline{M}ulti-\underline{A}gent \underline{R}epair \underline{S}earch method called \system that performs Monte Carlo Tree Search (MCTS) problem over the search space. Figure~\ref{fig: system overview} presents an overview of \system. Specifically, each node in the search tree represents a system state, as described in the previous section. \system continuously selects a repair operation that transforms a system state to another state until it reaches a search limit or a refined output that it deems satisfiable. While \system adopts the widely used Upper Confidence Bound for Trees (UCT) heuristic~\cite{coulom2006efficient} for node selection, we introduce two new designs that adapt the standard MCTS algorithm to efficient MAS repair with low token cost. First, to enable fine-grained control over the repair exploration process, we define an action space that distinguishes failure attribution from repair and also supports partial rollouts and continuation.  Second, to evaluate partially completed trajectories, we introduce a novel taxonomy-augmented evaluation mechanism that characterizes common failure patterns and generates dynamic reward signals for MCTS, thereby enabling informed search decisions before a complete trajectory becomes available.

\subsection{Partial Rollout and Action Space}
MCTS algorithms in domains such as code generation~\cite{zhang2023planning} and game playing~\cite{hao2023reasoning} commonly evaluate a node by rolling out to a terminal state and observing a verifiable outcome. However, for multi-agent systems, full trajectory rollouts are computationally expensive, since reaching a terminal state requires many additional agent turns and LLM calls. \system instead support partial rollouts, which  continues the execution of the multi-agent system for up to $L$ steps from the current state. This design bounds the cost of each rollout while allowing the search to build longer trajectories incrementally.


To support partial rollouts, we design two actions in the action space of \system,  \textit{Guided Repair} and \textit{Continuation}. Furthermore, we design a separate \textit{Rollback} action to decouple state restoration from repair execution to allow fine-grained control over the repair exploration process. This design is especially beneficial when the initial fault attribution is inaccurate, as it allows {\system} to perform multiple rollbacks without unnecessary re-execution.

\vspace{1pt}
\noindent \textbf{Rollback.}
Errors early in an MAS trajectory can propagate through later steps. Restarting the entire execution discards the valid prefix. \textit{Rollback} resolves this problem by truncating the trajectory at the faulty step $t$ and removing all subsequent steps, which preserves the valid prefix. \system then reloads the system state $s_t$  for a repair attempt.


\vspace{1pt}
\noindent \textbf{Guided Repair.} The \textit{Guided Repair} adds the fix guidance $\phi$ to the prompt for the next agent turn and resumes the MAS for $L$ steps. The guidance is a concrete suggestion for addressing a diagnosed failure cause (e.g., ``{\em Avoid the previous error X...}'') and corresponds to the feedback prompt shown in Figure~\ref{fig: system overview}. We explain how \system generates $\phi$ in the following subsection, \textit{Diagnosis-guided Node Expansion}. Limiting the repair to $L$ steps avoids the cost of a full trajectory rollout.

\vspace{1pt}
\noindent \textbf{Continuation.}
For \textit{Continuation}, \system resumes the MAS from the current state without adding repair guidance and executes it for $L$ additional steps. This action extends a promising partial trajectory and explores another possible path from the selected node.

\subsection{Taxonomy-Augmented Evaluation and Diagnosis}
While supporting partial rollouts reduces token costs, it introduces a new challenge as a partial trajectory has no final task outcome to evaluate on and the evolving nature of MAS interactions makes an intermediate node's contribution to the final solution difficult to assess. To address this challenge, we propose an evaluation and diagnosis mechanism that leverages known MAS failure patterns to identify early signs of failure in partial trajectories. Furthermore, we propose a granular scoring rubric to measure partial task progress. This mechanism is implemented using an LLM judge, whose complete prompt is provided in the appendix.

 
\vspace{1pt}
\noindent\textbf{MAS Failure  Recognition and Diagnosis.} We curated 14 failure patterns based on the MAST taxonomy, which was rigorously developed by a comprehensive analysis of 150 MAS failure trajectories~\citep{cemri2025multi}. Some example patterns include {\em step repetition}, {\em not following task specification}, {\em violating system role}, {\em ignoring input from other agents}, and {\em premature termination}. Each pattern is encoded as a checklist item with a concise definition in the evaluation prompt to the LLM judge. Guided by these failure patterns, the LLM judge is instructed to perform a comprehensive analysis of a given trajectory to identify any potential issues and infer likely root causes. We denote these results as diagnostic feedback $f$. This feedback informs the trajectory score $r$, which is assigned using the fine-grained rubric described below. The feedback $f$ is also stored in the corresponding node in the search tree and used to guide subsequent expansion if that node is selected again. If no issue is found, the trajectory is deemed promising if it is a partial trajectory or successful if it is a complete trajectory with the final outcome included.

\noindent \textbf{Granular Scoring Rubric.} Inspired by single-step simulation strategies~\cite{zhang2025webpilot}, the LLM judge assigns each trajectory a scalar score $r \in [0, 1]$ to reflect the task progress. 
The rubric defines five levels. \textit{Verified Completion} (1.0) is reserved for a complete trajectory that satisfies three conditions. First, the judge detects no failure. Second, the final answer is supported by data returned through web search, file reading, or code execution, which we call \emph{supporting evidence}. Third, the judge derives the same answer using this evidence, which provides \emph{independent answer verification}. \textit{Strong Progress} (0.8--0.9) indicates substantial progress without a detected failure, although completion has not been fully established. \textit{Moderate Progress} (0.5--0.7) indicates useful progress with important steps unresolved or progress stalled. \textit{Limited Progress} (0.2--0.4) indicates little progress or a clear but recoverable error. \textit{Failure / Divergence} (0.0--0.1) indicates severe errors, repeated loops, substantial task drift, or a hallucinated final answer. A partial trajectory is scored by its current progress and is not treated as a failure merely because it lacks a final answer or is short. \system then uses standard MCTS backpropagation~\cite{coulom2006efficient} to propagate $r$ through the search tree.

\subsection{Diagnosis-guided Node Expansion}

During expansion, an LLM-based action selector first examines the diagnostic feedback $f$ from the node's previous evaluation against its trajectory. This design is motivated by our observation that sometimes the LLM judge may misinterpret a partial trajectory. Therefore, the action selector performs an examination first before acting to detect any diagnosis errors, inspired by prior work on self-reflection~\cite{madaan2023self}. If the selector confirms the diagnosis of a failure, it uses the failure cause described in $f$ to determine the rollback step $t$ and generate a specific fix suggestion $\phi$. If it agrees that the partial trajectory contains no failure, the selector favors extending the current trajectory. When its own assessment differs from the stored diagnosis $f$, the selector can choose a different expansion action. Once the selector makes its decision, \system applies it to create a child node in the search tree. If \system later selects the same node again, the selector repeats this process to create another child. The node is fully expanded once its number of children reaches the maximum branching factor $K$. The full prompt for the action selector is described in the appendix.

\section{Benchmark} 
\label{sec: dataset}


We introduce \dataset, a large-scale benchmark for evaluating MAS repair methods. While benchmarks from existing MAS failure attribution work provide static failure logs and attribution annotations, they lack the execution states needed to replay and repair a trajectory. DoVer~\cite{ma2025dover} describes its failure trace collection procedure and releases code for regenerating the traces using an interactive MAS debugging tool~\cite{epperson2025interactive}. However, it does not preserve the intermediate system states needed to replay a specific failure trajectory. When reproducing failure traces, rerunning the agents does not guarantee the same traces due to the stochasticity of agent executions. Besides, DoVer collects failure traces using one agent architecture and one LLM backbone. In comparison, \dataset covers four agent architectures and four LLM backbones.  \dataset also records intermediate system states throughout each execution, allowing a repair method to restore and modify the trajectory from any saved state. We build \dataset with the Microsoft Agentic Framework (MAF)~\cite{microsoft2026agent}, an open-source framework for developing multi-agent workflows.


\dataset includes two key features. First, it provides APIs for deterministic replay of failure trajectories by setting the LLM temperature to zero and preserving all system configurations. Second, it can load the serialized MAS state at any execution step $t$ and resume execution from that point. 


\vspace{1pt}
\noindent \textbf{Benchmark Construction.} Following recent MAS failure attribution work~\cite{zhang2025agent,cemri2025multi} and DoVer~\cite{ma2025dover}, we run tasks from GAIA~\cite{mialon2023gaia} and AssistantBench~\cite{yoran2024assistantbench} validation set to collect failure trajectories. GAIA evaluates general assistants on real-world, multi-modal questions that require reasoning and tool use, including web navigation and file handling. AssistantBench evaluates assistants on realistic, time-intensive web navigation tasks.



To collect diverse failure trajectories, we construct MAS with four widely used agent architectures identified in recent surveys~\cite{li2024survey, luo2503large}, including {\em centralized}, {\em sequential}, {\em decentralized}, and {\em concurrent}.  Following Who\&When~\cite{zhang2025agent}, each architecture uses the five-agent team from Magentic-One~\cite{fourney2024magentic}. Furthermore, we use four LLM backbones, including Qwen3.5-9B, Qwen3-30B-A3B, Nemotron-3-Nano-30B-A3B, and GPT-5.4-mini. These models span different model families and sizes, including both open-source and closed-source models. Together, these choices yield 16 MAS configurations with varied communication structures and underlying models.

Running all 16 configurations on every task would incur substantial token costs and commercial API fees. We therefore run them on 66 GAIA tasks that can be solved using free tools and APIs, such as the Wikipedia API~\cite{wikipedia2026}. We call this partition \textit{GAIA-Free}. The remaining 99 GAIA tasks and all 33 AssistantBench validation tasks require paid services and APIs such as Firecrawl~\cite{firecrawl2026}. To control the cost, we run these tasks using the centralized architecture with each of the four LLM backbones. We call these two partitions \textit{GAIA-Comm} and \textit{AssistantBench}. Each execution records a trajectory and a replayable MAS state at every execution step. We label an execution as successful if the solution produced by the MAS matches the benchmark ground truth and as failed otherwise. In total, this process produces 1,584 initial trajectories, including 274 successful and 1,310 failed executions. Table~\ref{tab: benchmark_failure_counts} reports the distribution of the failed trajectories. Additional construction details are provided in the appendix.

\begin{table}[t]
    \centering
    \caption{The distribution of failure trajectories in \dataset}
    \label{tab: benchmark_failure_counts}
    \footnotesize
    \setlength{\tabcolsep}{3pt}
    \resizebox{\columnwidth}{!}{%
    \begin{tabular}{llrrrr}
        \toprule
        \textbf{Partition} & \textbf{Arch.}
        & \textbf{Qwen3} & \textbf{Nemontron3}
        & \textbf{Qwen3.5} & \textbf{GPT} \\
        \midrule
        \multirow{4}{*}{GAIA-Free}
        & Cent.   & 58 & 57 & 54 & 45 \\
        & Seq.    & 60 & 63 & 64 & 40 \\
        & Decent. & 62 & 63 & 63 & 49 \\
        & Concur. & 59 & 61 & 54 & 44 \\
        \midrule
        GAIA-Comm
        & Cent. & 83 & 84 & 72 & 61 \\
        AssistantBench
        & Cent. & 28 & 31 & 29 & 26 \\
        \bottomrule
    \end{tabular}
    }
\end{table}

\begin{table*}[t]
    \centering
    \caption{Post-repair pass rate across GAIA and AssistantBench benchmarks with centralized orchestration.}
    \label{tab: main_results_benchmark}
    \small
    \setlength{\tabcolsep}{4.2pt}
    \resizebox{0.85\textwidth}{!}{%
    \begin{tabular}{l ccccc ccccc}
        \toprule
        & \multicolumn{5}{c}{\textbf{GAIA}} & \multicolumn{5}{c}{\textbf{AssistantBench}} \\
        \cmidrule(lr){2-6} \cmidrule(lr){7-11}
        \textsc{Model}
        & $Initial$ & Reflexion & ReAct & DoVer & \system
        & $Initial$ & Reflexion & ReAct & DoVer & \system \\
        \midrule
        Qwen3.5-9B
        & 23.6\% & 29.7\% & 37.0\% & 35.2\% & \bestscore{44.8\%}
        & 12.1\% & 15.2\% & 27.3\% & 24.2\% & \bestscore{36.4\%} \\
        \addlinespace
        Qwen3-30B-A3B
        & 14.5\% & 17.6\% & 24.8\% & 23.6\% & \bestscore{33.9\%}
        & 15.2\% & 21.2\% & 27.3\% & 30.3\% & \bestscore{36.4\%} \\
        \addlinespace
        Nemotron-3-Nano-30B-A3B
        & 14.5\% & 17.0\% & 20.0\% & 22.4\% & \bestscore{32.1\%}
        & 6.1\% & 9.1\% & 18.2\% & 15.2\% & \bestscore{27.3\%} \\
        \addlinespace
        GPT-5.4-mini
        & 35.8\% & 43.6\% & 52.1\% & 51.5\% & \textbf{60.0\%}
        & 21.2\% & 27.3\% & 33.3\% & 33.3\% & \textbf{42.4\%} \\
        \bottomrule
    \end{tabular}
    }
\end{table*}

\begin{table*}[t]
    \centering
    \caption{Post-repair pass rate across four orchestrations on GAIA-Free.}
    \label{tab: main_results}
    \small
    \setlength{\tabcolsep}{10pt}
    \resizebox{0.85\textwidth}{!}{%
    \begin{tabular}{llccccc}
        \toprule
        & & & \multicolumn{4}{c}{\textbf{Repair Method}} \\
        \cmidrule(lr){4-7}
        \textsc{Model}
        & \textsc{Orchestration}
        & $Initial$
        & Reflexion
        & ReAct
        & DoVer
        & \system \\
        \midrule
        \multirow{4}{*}{Qwen3.5-9B}
        & Centralized & 18.2\% & 22.7\% & 27.3\% & 28.8\% & \bestscore{36.4\%} \\
        & Sequential & 3.0\% & 6.1\% & 18.2\% & N/A & \bestscore{28.8\%} \\
        & Decentralized & 4.5\% & 10.6\% & 22.7\% & N/A & \bestscore{34.8\%} \\
        & Concurrent & 18.2\% & 22.7\% & 28.8\% & N/A & \bestscore{34.8\%} \\
        \midrule
        \multirow{4}{*}{Qwen3-30B-A3B}
        & Centralized & 12.1\% & 15.2\% & 19.7\% & 21.2\% & \bestscore{30.3\%} \\
        & Sequential & 9.1\% & 10.6\% & 18.2\% & N/A & \bestscore{27.3\%} \\
        & Decentralized & 6.1\% & 10.6\% & 13.6\% & N/A & \bestscore{19.7\%} \\
        & Concurrent & 10.6\% & 18.2\% & 22.7\% & N/A & \bestscore{30.3\%} \\
        \midrule
        \multirow{4}{*}{Nemotron-3-Nano-30B-A3B}
        & Centralized & 13.6\% & 15.2\% & 16.7\% & 21.2\% & \bestscore{24.2\%} \\
        & Sequential & 4.5\% & 9.1\% & 15.2\% & N/A & \bestscore{19.7\%} \\
        & Decentralized & 4.5\% & 6.1\% & 10.6\% & N/A & \bestscore{13.6\%} \\
        & Concurrent & 7.6\% & 10.6\% & 13.6\% & N/A & \bestscore{19.7\%} \\
        \midrule
        \multirow{4}{*}{GPT-5.4-mini}
        & Centralized & 31.8\% & 40.9\% & 50.0\% & 47.0\% & \bestscore{57.6\%} \\
        & Sequential & 39.4\% & 43.9\% & 51.5\% & N/A & \bestscore{59.1\%} \\
        & Decentralized & 25.8\% & 39.4\% & 43.9\% & N/A & \bestscore{54.5\%} \\
        & Concurrent & 33.3\% & 40.9\% & 47.0\% & N/A & \bestscore{56.1\%} \\
        \bottomrule
    \end{tabular}
    }
\end{table*}


        




\section{Experiments}



\subsection{Experiment Setup}


\vspace{1pt}
\noindent \textbf{Baseline Methods.} We compare \system with DoVer~\cite{ma2025dover} and two repair baselines adapted from Reflexion~\cite{shinn2023reflexion} and ReAct~\cite{yao2023react}. DoVer is the state-of-the-art automated MAS repair method. It performs a linear search over failure-attribution hypotheses. Its repair interventions edit only the orchestrator's plans or messages to subagents, followed by a full rollout from each intervention point. The Reflexion-based baseline examines the current failed trajectory, generates a reflection that identifies likely errors and suggests improvements, and inserts this reflection-derived feedback at the latest resumable checkpoint, and continues execution linearly. The ReAct-based baseline repairs the trajectory through a sequential reasoning-and-action loop. At each iteration, it examines the current trajectory, reasons about the next intervention, selects the action, and observes the resulting execution before making its next decision. It uses the same action space as \system but follows a single repair path without tree search. We denote these two baselines as Reflexion and ReAct for short in the tables. We select them as baselines because they support iterative interaction with an executable environment and can therefore modify and rerun failed MAS trajectories.

\vspace{1pt}
\noindent \textbf{Metrics.} Following the outcome-oriented evaluation of DoVer~\cite{ma2025dover}, we report the initial pass rate and post-repair pass rate. The initial pass rate is the percentage of tasks solved by the MAS before repair. The post-repair pass rate is the percentage of tasks successfully solved after a repair method is applied to the initially failed executions.


\vspace{1pt}
\noindent \textbf{Implementation Details.} We use the same LLM backbone for repair as in the original execution and implement all methods with the Microsoft Agentic Framework (MAF)~\cite{microsoft2026agent}. DoVer is limited to centralized orchestration because it focuses on editing a central orchestrator's plans and messages to subagents. We therefore evaluate DoVer only under the centralized agent architecture and report N/A for the other architectures. For \system, we set branching factor $K$ to 3, partial-rollout length $L$ to 4 steps, maximum depth to 8, the UCT exploration weight to 1.41, and maximum search budget to 75 iterations. The LLM-as-a-judge and action selector use temperatures of 0.2 and 0.7, respectively.

 \begin{table*}[t]
      \centering
      \caption{Repair-time token consumption across four orchestration and four LLM backbones.}
      \label{tab: efficiency_analysis_comprehensive}
      \small
      \setlength{\tabcolsep}{8pt}
      \resizebox{0.8\textwidth}{!}{%
      \begin{tabular}{llrrrr}
          \toprule
          \textsc{Model} & \textsc{Method}
          & \textbf{Centralized}
          & \textbf{Sequential}
          & \textbf{Decentralized}
          & \textbf{Concurrent} \\
          \midrule
          \multirow{4}{*}{Qwen3.5-9B}
          & Reflexion & \textbf{1,916,430} & 3,088,399 & 2,748,577 & \textbf{2,953,488} \\
          & ReAct & 2,243,906 & \textbf{1,668,761} & \textbf{1,371,421} & 3,175,610 \\
          & DoVer & 2,075,860 & N/A & N/A & N/A \\
          & \system & 2,451,982 & 1,968,894 & 1,644,313 & 4,193,241 \\
          \midrule

          \multirow{4}{*}{Qwen3-30B-A3B}
          & Reflexion & 1,664,699 & 1,436,719 & 1,973,608 & 4,904,795 \\
          & ReAct & \textbf{682,517} & \textbf{405,198} & \textbf{763,907} & \textbf{1,305,062} \\
          & DoVer & 768,275 & N/A & N/A & N/A \\
          & \system & 861,915 & 658,318 & 806,733 & 2,091,065 \\
          \midrule

          \multirow{4}{*}{Nemotron-3-Nano-30B-A3B}
          & Reflexion & 841,352 & 1,392,854 & 2,436,846 & 3,514,361 \\
          & ReAct & \textbf{424,179} & 762,944 & \textbf{527,884} & \textbf{1,534,598} \\
          & DoVer & 611,048 & N/A & N/A & N/A \\
          & \system & 617,550 & \textbf{537,620} & 620,946 & 1,874,168 \\
          \midrule

          \multirow{4}{*}{GPT-5.4-mini}
          & Reflexion & 1,840,274 & 1,262,636 & 1,592,576 & 1,590,872 \\
          & ReAct & 1,239,844 & \textbf{1,067,292} & \textbf{952,362} & \textbf{1,390,122} \\
          & DoVer & \textbf{1,023,725} & N/A & N/A & N/A \\
          & \system & 1,455,252 & 1,300,835 & 1,130,827 & 1,685,436 \\
          \bottomrule
      \end{tabular}
      }
  \end{table*}

\subsection{Main Results}

\vspace{1pt}
\noindent \textbf{Effectiveness across Benchmarks.} Table~\ref{tab: main_results_benchmark} reports post-repair pass rates on all GAIA and AssistantBench validation tasks under the centralized orchestration. Across all four LLM backbones, \system achieves the highest post-repair pass rate on both benchmarks. Its improvement over the strongest baseline ranges from 7.8\% to 9.7\% on GAIA and from 6.1\% to 9.1\% on AssistantBench. For example, \system improves the GAIA result with Nemotron-3-Nano-30B-A3B from DoVer's 22.4\% to 32.1\%. On AssistantBench with Qwen3.5-9B, it improves the strongest baseline from 27.3\% to 36.4\%. These results show that the repair gains hold across both benchmarks and all four LLM backbones.


\vspace{1pt}
\noindent \textbf{Effectiveness across MAS Architectures.} Table~\ref{tab: main_results} reports the effectiveness of repairing MAS with four different architectures. \system achieves the highest post-repair pass rate under every MAS architecture and LLM backbone. Its improvement over the strongest applicable baseline ranges from 3.0\% to 12.1\%. The largest gain occurs with Qwen3.5-9B under the decentralized architecture, where \system reaches 34.8\% compared with ReAct's 22.7\%. The gains also hold for GPT-5.4-mini, ranging from 7.6\% under the centralized and sequential orchestrations to 10.6\% under the decentralized orchestration.

\begin{table}[t]
  \centering
  \caption{Ablation study on GAIA-Free with Qwen3-30B-A3B under the centralized orchestration.}
  \label{tab: ablation_results}
  \footnotesize
  \setlength{\tabcolsep}{6pt}
  \begin{tabular}{lrr}
      \toprule
      \textbf{Variant} & \textbf{ Pass Rate} & \textbf{Tokens} \\
      \midrule
      Linear Search ($K=1$) & \textit{22.7\%} & \textit{824,591} \\
      Full Rollout & \textit{22.7\%} & \textit{3,140,488} \\
      \system w/o Tax. & 25.8\% & \textit{714,680} \\
      \system (Full) & \textbf{30.3\%} & \textit{861,915} \\
      \bottomrule
  \end{tabular}
\end{table}

\subsection{Token Cost Analysis}

Table~\ref{tab: efficiency_analysis_comprehensive} reports total repair-time token consumption across four LLM backbones and four MAS architectures. Each value reports the average number of tokens consumed to repair one failed trajectory, including both repair decisions and the resulting MAS execution. Compared with Reflexion, \system reduces token consumption by 1.6\% to 54.7\%. Reflexion accumulates reflection and interaction context across repair attempts, whereas the \textit{Rollback} action in \system removes an erroneous trajectory suffix before execution continues. This difference allows \system to explore more repair alternatives without retaining the complete history of every attempt.

Compared with ReAct, \system incurs a relative increase in token cost from 14.0\% to 38.7\% across the four LLM backbones. Its relative performance improvement over ReAct ranges from 56.2\% to 92.5\%, which is greater than the relative token cost overhead. 
Furthermore, compared to DoVer, \system's relative cost increase ranges from 1.1\% to 42.2\%, while its relative performance improvement ranges from 39.5\% to 100\%. Thus, the relative improvement in repair effectiveness exceeds the relative increase in token consumption for both comparisons.

\subsection{Ablation Study}

We conduct the ablation study on GAIA-Free using Qwen3-30B-A3B under centralized orchestration. We evaluate the three main design choices in \system: tree search, partial rollouts, and taxonomy-guided evaluation. \textit{Linear Search} sets the branching factor to $K=1$, producing a single repair path while retaining the other components. \textit{Full Rollout} uses an action space consisting of \textit{Rollback} and \textit{Guided Repair}. In this variant, \textit{Guided Repair} performs a full rollout from the selected checkpoint to MAS termination. \textit{\system w/o Tax.} removes the MAS failure patterns from the evaluation prompt but retains the granular scoring rubric. All other experimental settings remain unchanged.

Table~\ref{tab: ablation_results} shows that the full system achieves the highest post-repair pass rate of 30.3\%. Linear search reduces the pass rate from 30.3\% to 22.7\% because $K=1$ prevents the search from comparing alternative repair paths. The full-rollout variant also achieves 22.7\% but increases token consumption from 861,915 to 3,140,488. Full rollouts consume more tokens and delay reevaluation, allowing errors to propagate before the search can redirect the trajectory. Removing the taxonomy reduces the pass rate from 30.3\% to 25.8\% because the judge lacks explicit failure patterns for diagnosing the trajectory and guiding later repair decisions.

\subsection{Stochasticity Analysis}
We measure run-to-run variation by repeating \system three times for each orchestration on GAIA-Free with Qwen3-30B-A3B. All runs use the same initial trajectories and hyperparameters. Table~\ref{tab: stochasticity} reports the minimum, maximum, and mean post-repair pass rates. The difference between the minimum and maximum is 1.5\% for the sequential and decentralized orchestrations and 3.0\% for the centralized and concurrent orchestrations. The mean differs from the corresponding result in main results table by at most 1.5\%. This limited variation indicates that the main results are not dominated by a single outlier run.

\subsection{Sensitivity Analysis}
We vary the branching factor $K$, iteration budget, and generation steps $L$ on GAIA-Free using Qwen3-30B-A3B under centralized orchestration. Performance initially improves as $K$ increases, then decreases since wider branching reduces search depth. Increasing the iteration budget consistently improves performance, although the gains diminish at higher budget. Moderate rollout lengths outperform both short rollouts that limit progress and long rollouts that delay reevaluation. Full results are provided in the appendix.

\begin{table}[t]
    \centering
    \caption{Stochasticity results with Qwen3-30B-A3B.}
    \label{tab: stochasticity}
    \footnotesize
    \setlength{\tabcolsep}{4pt}
    \resizebox{0.75\linewidth}{!}{%
    \begin{tabular}{lccc}
        \toprule
        \multirow{2}{*}{\textbf{Orchestration}}
        & \multicolumn{3}{c}{\textbf{Post-Repair Pass Rate}} \\
        \cmidrule(lr){2-4}
        & \textbf{Minimum} & \textbf{Maximum} & \textbf{Mean} \\
        \midrule
        Centralized   & 27.3\% & 30.3\% & 28.8\% \\
        Sequential    & 25.8\% & 27.3\% & 26.8\% \\
        Decentralized & 18.2\% & 19.7\% & 18.7\% \\
        Concurrent    & 30.3\% & 33.3\% & 31.8\% \\
        \bottomrule
    \end{tabular}
    }
\end{table}

\section{Related Work}

\vspace{1pt}
\noindent \textbf{Failure Attribution in Multi-Agent Systems.} Recent work studies which agent and step cause an MAS task failure. MAST~\cite{cemri2025multi} catalogs failure patterns involving task specification, inter-agent coordination, and verification. Who\&When~\cite{zhang2025agent} evaluates all-at-once, step-by-step, and binary-search attribution methods for locating the responsible agent and decisive error step from execution logs. AgenTracer~\cite{zhang2026agentracer} trains a specialized failure tracer on automatically constructed trajectories. RAFFLES~\cite{zhu2026raffles} uses a central LLM judge to investigate faults and specialized evaluators to assess both candidate faults and the judge's reasoning. AgentForesight~\cite{zhang2026agentforesight} instead audits partial trajectories online to flag decisive errors during execution. These methods identify or predict failures but do not execute repairs. In contrast, our work uses failure diagnosis to guide a search over executable repairs.

\vspace{1pt}
\noindent \textbf{Repair for Multi-Agent Systems.} Interactive debugging tools such as AGDebugger~\cite{epperson2025interactive} and LangGraph~\cite{langgraphstudio2024} support checkpointing, rollback, editing, and re-execution. However, users must choose and validate each intervention. General agent methods such as Reflexion~\cite{shinn2023reflexion} and ReAct~\cite{yao2023react} improve agent execution through feedback or interleaved reasoning and action, but do not search over stateful MAS repair alternatives. To the best of our knowledge, DoVer~\cite{ma2025dover} is the only method that is specifically designed for automated MAS repair. Given a failure trajectory, DoVer proposes failure attribution hypotheses and converts the hypotheses to repair edits to the orchestrator's plans or messages to sub-agents. Then, it performs a full rollout from each intervention point. This sequential process incurs high token costs. Its orchestrator-level edits also restrict it to centralized agent architectures. In contrast, \system formulates MAS repair as a Monte Carlo Tree Search problem, enabling efficient exploration of a much larger repair space through partial rollouts that substantially reduce token costs. \system also supports repair edits to any point in the trajectory, rather than restricting edits to the orchestrator alone. 



\vspace{1pt}
\noindent \textbf{Benchmarks for MAS Repair.} Existing failure-attribution benchmarks~\cite{zhang2025agent, cemri2025multi, zhang2026agentforesight, zhang2026agentracer} primarily store execution logs and failure annotations. They do not preserve the runtime states required to restore and modify an MAS execution. DoVer~\cite{ma2025dover} performs repair within live debugging sessions, so it cannot release the resulting executions as a standalone, reusable benchmark. \dataset addresses this gap by storing replayable MAS states at every step across four agent architectures and four LLM backbones.

\vspace{1pt}
\noindent\textbf{Monte Carlo Tree Search for LLMs.} MCTS~\cite{coulom2006efficient} builds a search tree through repeated node selection, expansion, evaluation, and value backpropagation. Recent LLM work applies MCTS to various domains. PG-TD uses MCTS-guided planning to generate code~\cite{zhang2023planning}. GIF-MCTS generates and iteratively refines executable code world models~\cite{dainese2024generating}. Alpha-SQL searches over partial SQL constructions~\cite{li2025alpha}. SWE-Search combines MCTS with iterative refinement for repository-level software engineering tasks~\cite{antoniades2025swe}. Despite these advances, MCTS has not yet been applied to automated MAS repair. \system fills this gap by adapting the MCTS algorithm to support partial rollouts and specific actions for MAS repair.

\section{Conclusion}

We presented \system, an automated framework that formulates MAS repair as Monte Carlo Tree Search process. It searches for candidate repairs through partial rollouts, diagnosis-guided node expansion, and taxonomy-augmented evaluation. We also introduced \dataset, the first large-scale benchmark with 1,310 replayable MAS failure trajectories across four agent architectures and four LLM backbones. Compared to the state-of-the-art method, \system achieves an absolute performance improvement ranging from 8.5\% to 10.3\% on GAIA and from 6.1\% to 12.2\% on AssistantBench. Furthermore, \system consistently outperforms the strongest applicable baseline with an absolute improvement from 3.0\% to 12.1\%, when repairing failures from MAS with  different architectures. Ablation results show that the specialized repair actions and taxonomy-augmented evaluation work together to improve repair performance.

\newpage

\appendix

\section{Appendix}
\subsection{StateMAS Dataset}

Who\&When~\cite{zhang2025agent} includes 127 tasks selected from GAIA and AssistantBench. In contrast, \dataset covers the complete validation sets of both benchmarks and contains 1,584 trajectories. We execute each task under its assigned model-architecture configurations. Of these trajectories, 274 produce the correct answer without repair, while 1,310 do not and are subsequently processed by the repair methods. We label a trajectory \textit{Resolved} if it produces the correct final answer and \textit{Unresolved} otherwise.
 
We divide \dataset into three partitions. \textit{GAIA-Free} contains 66 tasks that can be solved using open-access resources, such as the Wikipedia API~\cite{wikipedia2026}, yielding 1,056 trajectories across four agent architectures and four LLM backbones. \textit{GAIA-Comm} contains the remaining 99 GAIA tasks, which require paid web-browsing APIs such as Firecrawl~\cite{firecrawl2026}, yielding 396 trajectories across four LLM backbones under the centralized architecture. \textit{AssistantBench} contains all 33 AssistantBench validation tasks, which involve realistic and time-intensive web navigation, yielding 132 trajectories under the same centralized setting.

\vspace{1pt}
\noindent \textbf{Base Models.} We evaluate \system with four LLM backbones spanning different sizes and architectures. We access the closed-source GPT-5.4-mini model through its official API. The three open-source models are Qwen3.5-9B, a dense Transformer; Qwen3-30B-A3B, which uses a Mixture-of-Experts architecture~\cite{shazeer2017outrageously}; and Nemotron-3-Nano-30B-A3B, which uses the Mamba architecture~\cite{gu2024mamba}. We serve the open-source models locally with vLLM~\cite{kwon2023efficient} using their default inference configurations.

\vspace{1pt}
\noindent \textbf{Agentic Systems.} We construct \dataset using four agent architectures derived from collaboration patterns identified in recent surveys~\cite{li2024survey, luo2503large}. Following Who\&When~\cite{zhang2025agent}, each architecture uses the five-agent team from Magentic-One~\cite{fourney2024magentic}. The agents specialize in capabilities such as web browsing and local file navigation. Using the same team across architectures isolates the effect of their communication and coordination structures.

\begin{enumerate}
    \item \textit{Sequential:} Agents operate as a pipeline. Each agent processes the task in turn and passes its output to the next agent. Systems such as MetaGPT~\cite{hong2023metagpt}, ChatDev~\cite{qian2024chatdev}, and OpenManus~\cite{openmanus2024} use this architecture for multistage workflows in which each step depends on previous outputs.

    \item \textit{Concurrent:} Multiple agents process the same task in parallel, and their outputs are subsequently aggregated. HyperAgent~\cite{phan2024hyperagent} adopts this architecture to collect diverse candidate solutions or perspectives.

    \item \textit{Centralized:} A central manager selects agents and coordinates the communication flow. Magentic-One~\cite{fourney2024magentic} and AppWorld~\cite{trivedi2024appworld} use this architecture to support iterative collaboration among specialized agents.

    \item \textit{Decentralized:} Agents communicate through direct handoffs without a central manager. Following the pattern introduced by OpenAI Swarm~\cite{openai2024swarm}, each agent decides when and to whom it should delegate the task.
\end{enumerate}

\vspace{1pt}
\noindent \textbf{Data Representation.} Evaluating MAS repair requires replayable execution states rather than static interaction logs. For a trajectory containing $T$ execution steps, \dataset stores a corresponding MAS state $s_t$ after every step $t$. Each state captures the accumulated context, agent memories, and runtime variables at that point. We set the LLM temperature to zero and preserve the system configuration to support deterministic replay. A repair method can reload $s_t$ to reproduce the remaining execution or modify the restored state before resuming it. These operations enable counterfactual evaluation of how an intervention changes the final outcome.

Each trajectory in \dataset contains four components:
\begin{enumerate}
    \item \textit{Query}: The original task prompt drawn from GAIA or AssistantBench.

    \item \textit{Execution Log}: The complete interaction trace. For a failed execution, this trace provides the diagnostic input for repair.

    \item \textit{State Snapshots}: Serialized MAS states recorded after each execution step. These checkpoints allow a repair method to restore and modify the execution at a selected step.

    \item \textit{Workspace}: The task input files and intermediate artifacts produced during execution. These files preserve the context required to resume the MAS accurately.
\end{enumerate}

\begin{algorithm}[t]
\caption{Multi-Agent Repair via \system}
\label{alg: system_repair}
\begin{algorithmic}[1]
\INPUT $\tau_{fail}$, $N_{iter}$, $K$, $L$, $D_{max}$
\OUTPUT Highest-scoring trajectory $\hat{\tau}$

\STATE $root \leftarrow \text{Node}(\tau_{fail})$
\STATE $best\_traj \leftarrow \tau_{fail}$; \quad $max\_r \leftarrow 0$

\FOR{$i \leftarrow 1$ \textbf{to} $N_{iter}$}
    \STATE \textcolor{blue}{\COMMENT{\textbf{Phase 1: Selection}}}
    \STATE $node \leftarrow \text{SelectExpandable}(root,K,D_{max},\text{UCT})$

    \STATE \textcolor{blue}{\COMMENT{\textbf{Phase 2: Expansion}}}     
    \STATE $action \leftarrow \text{ActionSelector}(node)$
    \STATE $child \leftarrow \text{ApplyAction}(node, action, L)$
    
    \STATE \textcolor{blue}{\COMMENT{\textbf{Phase 3: Evaluation}}}
    \STATE $(r,f) \leftarrow \text{LLM\_Judge}(child.state)$
    \STATE $child.feedback \leftarrow f$
    
    \IF{$r > max\_r$} 
        \STATE $max\_r \leftarrow r$; \quad $best\_traj \leftarrow child.traj$
    \ENDIF
    
    \STATE \textcolor{blue}{\COMMENT{\textbf{Phase 4: Backpropagation}}}
    \STATE $curr \leftarrow child$
    \WHILE{$curr \neq \text{NULL}$}
        \STATE $N(curr) \leftarrow N(curr) + 1$
        \STATE $Q(curr) \leftarrow Q(curr) + r$
        \STATE $curr \leftarrow curr.parent$
    \ENDWHILE
    \IF{$r = 1$}
        \STATE \textbf{break}
    \ENDIF
\ENDFOR
\STATE \textbf{return} $best\_traj$
\end{algorithmic}
\end{algorithm}

\subsection{Detailed \system Algorithm}

Algorithm~\ref{alg: system_repair} presents the complete \system procedure. A node becomes fully expanded once it accumulates $K$ children across repeated visits. Each expansion calls the LLM-based action selector once and adds one child. \textsc{SelectExpandable} traverses fully expanded nodes from the root by selecting the child $c$ with the highest UCT value:
\begin{equation}
\operatorname{UCT}(c)=\frac{Q(c)}{N(c)}
 + C\sqrt{\frac{\ln N(p)}{N(c)}},
\end{equation}
where $p$ denotes the parent of $c$, $Q(c)$ is its cumulative score, $N(c)$ is its visit count, and $C$ controls exploration. Nodes at the maximum depth $D_{\max}$ cannot be expanded further.

\textsc{ApplyAction} creates a child using \textit{Rollback}, \textit{Guided Repair}, or \textit{Continuation}. The two generation actions execute the MAS for $L$ rounds during child construction, so \system requires no separate simulation rollout. The LLM judge then assigns the child a trajectory score $r$ and diagnostic feedback $f$. Standard MCTS backpropagation updates the selected search path using $r$.

The search terminates when the judge assigns \textit{Verified Completion} ($r=1$) or the search reaches its budget of $N_{\mathrm{iter}}$ expansions. \system returns the evaluated trajectory with the highest score. If the budget is exhausted and the returned trajectory does not produce a correct final answer, the repair attempt is counted as unsuccessful.




\begin{figure}[t]
    \centering
    \includegraphics[width=0.33\textwidth]{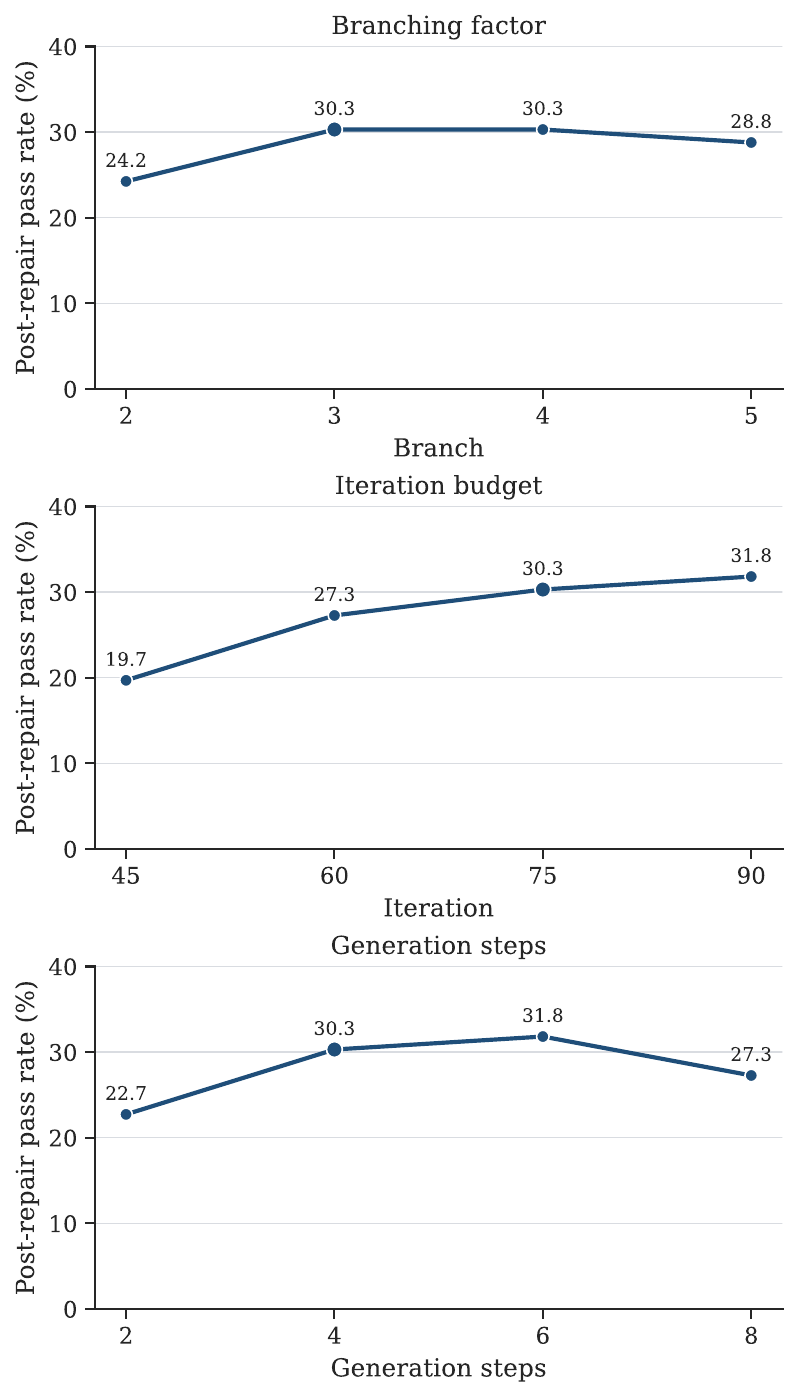}
    \caption{Sensitivity results with centralized orchestration}
    \label{fig: sensitivity}
\end{figure}

\subsection{Sensitivity Experiment}

We vary the branching factor $K$, MCTS iteration budget, and number of generation steps $L$ while holding the other settings fixed. All experiments use Qwen3-30B-A3B on GAIA-Free under centralized orchestration.

\vspace{1pt}
\noindent \textbf{Branching Factor.} Figure~\ref{fig: sensitivity} shows that increasing the branching factor from two to three raises the post-repair pass rate from 24.2\% to 30.3\%. A small branching factor limits the alternative \textit{Rollback}, \textit{Guided Repair}, and \textit{Continuation} branches that \system can explore. Performance remains at 30.3\% with four branches and decreases to 28.8\% with five. Under a fixed iteration budget, excessive branching expands the tree too broadly and leaves too few iterations to explore promising repair trajectories in depth.

\vspace{1pt}
\noindent \textbf{Iteration Budget.} The post-repair pass rate increases from 19.7\% with 45 iterations to 27.3\% with 60, 30.3\% with 75, and 31.8\% with 90. Additional iterations allow \system to evaluate more candidate branches, compare alternative repairs, and revisit promising trajectories. The improvement becomes smaller after 75 iterations, suggesting diminishing returns as the most promising branches receive more thorough exploration.

\vspace{1pt}
\noindent \textbf{Generation Steps.} The post-repair pass rate increases from 22.7\% with two generation steps to 30.3\% with four and 31.8\% with six. Short rollouts may end before the agents can make sufficient progress or fully apply the fix guidance. Increasing the rollout to eight steps reduces the rate to 27.3\%. Longer rollouts delay the next evaluation and expansion decision, which can allow errors to propagate before the search redirects the trajectory. 

\subsection{Limitation}
Our work has three limitations. First, we build \dataset with the Microsoft Agentic Framework, so its checkpoints and replay procedures may require adaptation for MAS implemented with other frameworks. Second, the benchmark tasks come from GAIA and AssistantBench and primarily cover general-assistant and web-navigation scenarios. Evaluation on domain-specific tasks is needed to establish broader generalizability. Third, computational and commercial API costs prevented us from evaluating all 16 combinations of four agent architectures and four LLM backbones on every task. We evaluate all 16 combinations on GAIA-Free, but use only the centralized architecture with four backbones on GAIA-Comm and AssistantBench.

\subsection{Prompt Templates}

\begin{tcolorbox}[
    enhanced,
    breakable,
    colback=white,
    colframe=black,
    colbacktitle=black!5,
    coltitle=black,
    boxrule=0.5pt,
    arc=0pt,
    title={Taxonomy-Augmented Evaluation Prompt},
    fonttitle=\bfseries
]
\small
You are an expert evaluator for a Multi-Agent System (MAS). Your task is to assess the quality and health of the conversation trajectory.

\medskip
\noindent You must evaluate the trajectory along two dimensions:
\begin{enumerate}
    \item \textbf{Strategic Validity:} Are the current sub-goals logical and necessary for solving the Original User Query?
    \item \textbf{Execution Progress:} Have the agents effectively executed these sub-goals?
\end{enumerate}

\noindent\textbf{[MAST Taxonomy -- Failure Mode Checklist]}

\noindent Carefully screen the trajectory for the following specific failure patterns:

\medskip
\noindent\textbf{Category 1: Agent Execution Issues}
\begin{itemize}
    \item \textbf{FM 1.1 Disobey Task Spec:} Failure to adhere to specific constraints or requirements of a given task, leading to suboptimal outcomes.
    \item \textbf{FM 1.2 Disobey Role Spec:} Acting outside defined responsibilities (e.g., an analyst behaving like a coder).
    \item \textbf{FM 1.3 Step Repetition:} Unnecessary reiteration of previously completed steps without new context.
    \item \textbf{FM 1.4 Context Loss:} Disregarding recent history or corrections and reverting to an older state.
    \item \textbf{FM 1.5 Unaware of Termination:} Failing to recognize when termination criteria are met, causing unnecessary loops.
\end{itemize}

\noindent\textbf{Category 2: Inter-Agent Misalignment}
\begin{itemize}
    \item \textbf{FM 2.1 Conversation Reset:} Unwarranted restart of the dialogue, losing progress.
    \item \textbf{FM 2.2 No Clarification:} Failing to ask for help when data is unclear or incomplete.
    \item \textbf{FM 2.3 Task Derailment:} Deviating from the intended objective to irrelevant topics.
    \item \textbf{FM 2.4 Info Withholding:} Failing to share critical insights that other agents need.
    \item \textbf{FM 2.5 Ignored Input:} Disregarding corrections or output provided by peer agents.
    \item \textbf{FM 2.6 Reasoning-Action Mismatch:} Taking an action that contradicts the internal reasoning log.
\end{itemize}

\noindent\textbf{Category 3: Verification Issues}
\begin{itemize}
    \item \textbf{FM 3.1 Premature Termination:} Stopping before all objectives are visibly met.
    \item \textbf{FM 3.2 Incomplete Verification:} Partial or shallow checking (e.g., ``it runs'' rather than ``it works as requested'').
    \item \textbf{FM 3.3 Incorrect Verification:} False positives or failure to cross-check crucial information.
\end{itemize}

\medskip
\noindent\textbf{[Supporting Evidence Requirement]}

\noindent\textbf{CRITICAL:} Only set \texttt{found\_solution} to true if all of these conditions are met:
\begin{enumerate}
    \item A tool call, such as web search, file read, or code execution, returned actual data.
    \item The answer is directly derived from that tool output rather than from the agent's prior knowledge or reasoning alone.
    \item The specific data supporting the answer is visible in the tool results.
\end{enumerate}

\noindent If an agent states an answer without any tool providing the supporting data, this is hallucination. Set \texttt{found\_solution=false}, assign a score no greater than 0.3, and detect FM 3.3 (\textit{Incorrect Verification}).

\noindent If tool calls failed, returned errors, or returned empty results, the agents cannot have a valid answer. The score must be no greater than 0.3.

\noindent Be skeptical: agents frequently fabricate plausible-sounding answers. Verify that the claimed answer appears in a tool result.

\noindent If the agent says ``data unavailable,'' ``unable to find,'' ``not found,'' or gives a similar non-answer, this is not a solution. Set \texttt{found\_solution=false} and assign a score no greater than 0.3.

\medskip
\noindent\textbf{[Independent Answer Verification]}

\noindent After checking the tool evidence, verify the answer logic:
\begin{enumerate}
    \item Does the extracted answer directly and precisely answer the original question?
    \item Is the answer in the correct format? For example, if the query asks for a number, is the answer a number?
    \item Does the answer have the correct units and scale? For example, distinguish 17 from 17,000 when the source reports values in thousands.
    \item Is the answer complete? For example, if the query asks for a full name, do not accept only the first name.
    \item Could the answer refer to a different entity with a similar name?
\end{enumerate}

\noindent If the answer fails any of these checks, assign a score no greater than 0.7 and explain the issue.

\medskip
\noindent\textbf{[Answer Re-Derivation]}

\noindent After extracting the answer, perform this check:
\begin{enumerate}
    \item Find the specific tool output, such as a search result, file content, or code output, on which the answer is based.
    \item Quote the exact text from that tool output that supports the answer.
    \item Independently derive the answer using only that quoted text.
    \item If the independent derivation differs from the agent's answer, the agent likely made an error. Assign a score no greater than 0.5 and explain the discrepancy.
    \item If no tool output directly supports the answer, treat the answer as hallucinated and assign a score no greater than 0.3.
\end{enumerate}

\medskip
\noindent\textbf{[Evaluation Rubric]}
\begin{itemize}
    \item \textbf{1.0 (Verified Completion):} The answer is directly supported by quoted tool evidence, the re-derivation matches, and no failure modes are detected.
    \item \textbf{0.8--0.9 (Strong Progress):} The trajectory makes substantial progress with logical sub-goals and no detected failure. If an answer is present, it is supported by tool evidence with at most minor ambiguity.
    \item \textbf{0.5--0.7 (Moderate Progress):} The trajectory makes useful progress but has unresolved steps or shows stagnation. If an answer is present, its evidence may be weak or indirect, or the re-derivation may produce a different result.
    \item \textbf{0.2--0.4 (Limited Progress):} The trajectory makes little progress or contains a clear but recoverable error, such as an unsupported answer or incorrect units, entity, or scale.
    \item \textbf{0.0--0.1 (Failure/Divergence):} The trajectory contains severe errors, repeated loops, substantial task drift, or a hallucinated terminal answer. A complete trajectory that terminates without an answer also falls within this level.
\end{itemize}

\medskip
\noindent\textbf{[Fresh-Start Awareness]}

\noindent If the trajectory is very short, it was likely recently truncated to remove an earlier error. Do not penalize short trajectories without failure mode. 

\medskip
\noindent\textbf{[Output Format]}

\texttt{Found\_solution: true OR false}\\
\noindent\texttt{Detected\_failure\_mode: "None" OR "FM X.Y" OR "Other"}\\
\texttt{Reasoning: [Explain why the direction is valid or invalid and cite specific interactions.]}\\
\texttt{Score: [Float between 0.0 and 1.0]}

\medskip
\noindent\textbf{[Input Context]} Analyze the following trajectory and assign a score using the evaluation rubric.

\medskip
\noindent\textbf{[Original User Query]}\\
\texttt{\{user\_query\}}

\medskip
\noindent\textbf{[Full Trajectory History]}\\
\texttt{\{trajectory\}}
\end{tcolorbox}

\begin{tcolorbox}[
    enhanced,
    breakable,
    colback=white,
    colframe=black,
    colbacktitle=black!5,
    coltitle=black,
    boxrule=0.5pt,
    arc=0pt,
    title={Feedback-Guided Expansion Prompt},
    fonttitle=\bfseries
]
\small
You are the Decision-Making Module for a Multi-Agent System repair process.

\medskip
\noindent You will be given:
\begin{enumerate}
    \item The full Trajectory History of the multi-agent conversation.
    \item Diagnostic Feedback from the previous evaluation phase, which includes a detected failure mode (if any), reasoning, and a numerical score.
\end{enumerate}

\noindent Your task is to decide the next action for the search tree by following these steps:

\medskip
\noindent\textbf{[Step 1: Verify]}

Cross-check the Diagnostic Feedback against the actual Trajectory History. Does the feedback accurately describe what happened? Is the identified failure mode (if any) consistent with the trajectory?

\medskip
\noindent\textbf{[Step 2: Decide]}

\noindent Based on your verification:
\begin{itemize}
    \item If you agree with the failure assessment (a real failure mode was detected and confirmed), choose \textit{rollback}. Identify the earliest checkpoint where the failure began and provide concrete fix guidance.
    \item Otherwise, choose \textit{guided repair} if stored fix guidance is available from a prior rollback; if
  not, choose \textit{continuation}.
\end{itemize}

\noindent\textbf{Available Actions:}
\begin{enumerate}
    \item Rollback: Roll back to the decisive error checkpoint and provide fix guidance for a new attempt.
    \item Guided Repair: Continue the agent conversation from the restored state while explicitly conditioning it on the stored fix guidance from rollback.
    \item Continuation: Continue the agent conversation from the current state.
\end{enumerate}

\medskip
\noindent\textbf{Constraint:} You must call exactly one function ((\textit{rollback}, \textit{guided repair}, or \textit{continuation}) exactly once. Do not call multiple functions. Your entire response must be a single function invocation.

\medskip
\noindent\textbf{[Trajectory History]}\\
\texttt{\{trajectory\}}

\medskip
\noindent\textbf{[Diagnostic Feedback from Previous Evaluation]}\\
\texttt{\{diagnostic\_feedback\}}
\end{tcolorbox}

\bibliography{aaai2027}

\end{document}